\documentclass[11pt,a4paper]{article}

\usepackage{authblk}
\usepackage[hyperref]{eacl2021}
\usepackage{times}
\usepackage{latexsym}
\usepackage{graphicx}
\usepackage{amsmath}
\usepackage{amssymb}

\usepackage{textcomp}  

\usepackage{booktabs}

\usepackage{microtype}

\aclfinalcopy 

\title{Scaling Federated Learning for Fine-tuning of Large Language Models}

\author[*]{Agrin Hilmkil}
\author[*]{Sebastian Callh}
\author[*]{Matteo Barbieri}
\author[+]{Leon René Sütfeld}
\author[+]{Edvin Listo Zec}
\author[+]{Olof Mogren}

\affil[*]{Peltarion, \href{https://peltarion.com}{peltarion.com}}
\affil[+]{RISE Research Institutes of Sweden, \href{https://ri.se}{ri.se}}

\date{2020-02-01}

\begin{document}
\maketitle
\begin{abstract}

Federated learning (FL) is a promising approach to distributed compute, as well as distributed data, and provides a level of privacy and compliance to legal frameworks. This makes FL attractive for both consumer and healthcare applications. While the area is actively being explored, few studies have examined FL in the context of larger language models and there is a lack of comprehensive reviews of robustness across tasks, architectures, numbers of clients, and other relevant factors. In this paper, we explore the fine-tuning of Transformer-based language models in a federated learning setting. We evaluate three popular BERT-variants of different sizes (BERT, ALBERT, and DistilBERT) on a number of text classification tasks such as sentiment analysis and author identification. We perform an extensive sweep over the number of clients, ranging up to 32, to evaluate the impact of distributed compute on task performance in the federated averaging setting. While our findings suggest that the large sizes of the evaluated models are not generally prohibitive to federated training, we found that the different models handle federated averaging to a varying degree. Most notably, DistilBERT converges significantly slower with larger numbers of clients, and under some circumstances, even collapses to chance level performance. Investigating this issue presents an interesting perspective for future research. 

\end{abstract}

\section{Introduction}

Transformer-based architectures such as BERT have recently lead to breakthroughs in a variety of language-related tasks, such as document classification, sentiment analysis, question answering, and various forms of text-mining \citep{vaswani2017attention, devlin2019bert, adhikari2019docbert, sun2019utilizing, yang2019end, lee2020biobert}. These models create semantic representations of text, which can subsequently be used in many downstream tasks \citep{devlin2019bert}. The training process for Transformers typically includes two phases: During pre-training, the model learns to extract semantic representations from large, task-independent corpora. The pre-training is followed by task-specific fine-tuning on a separate dataset to optimize model performance further.

In this paper, we study the effects of fine-tuning Transformer-based architectures in a federated learning (FL) setting. In FL, models are trained in a decentralized fashion on a number of local compute instances, called clients, and intermittently aggregated and synchronized via a central server. As such, FL is a solution for distributed compute, as well as distributed data, and provides a level of privacy with regards to the sharing of personal or otherwise sensitive data. Model aggregation is commonly performed via averaging of the weights of the individual client models, called Federated Averaging (\textsc{FedAvg}) \cite{mcmahan2017federated}.

Depending on the application, the number of clients in an FL setting can differ wildly. In instances where smartphones are used as clients, their number can reach into the millions \citep{hard2018federated}, whereas settings with higher compute requirements and more data per client will often range between a handful and a few dozens of clients. Here, we focus on the latter, as training large language models requires a lot of compute. A potential application of this is the medical field, in which automated analyses of electronic health records yield enormous potential for diagnostics and treatment-related insights \citep{zeng2018natural}.

\paragraph{Our contribution:} We provide a comprehensive overview of the applicability of the federated learning setting to large language models. To this end, we work with a fixed computation budget for each task, and use a fixed total amount of data while varying the number of clients between which the data is split up. This way, we isolate the effects of distributing data over several clients for distributed compute. We leave comparisons with a fixed amount of data per client and varying non-i.i.d. data distributions between clients for future work. The main contributions of this paper are the following: (1) We provide a comparison of three popular Transformer-based language models in the federated learning setting, using the IMDB, Yelp~F, and AG News datasets. (2) We analyze how the number of clients impacts task performance across tasks and model architectures.

\section{Related work}
Federated optimization was first introduced by \cite{konevcny2015federated}. The key challenges in this paradigm are communication efficiency when learning from many clients, privacy concerns with respect to leakage of client data, and variability in data distributions between clients (non-i.i.d. setting). \textsc{FedAvg} \cite{mcmahan2017federated} solves the federated optimization problem by building a global model based on local stochastic gradient descent updates and has been shown to work on non-i.i.d. data. Since then, many adaptations have arisen \cite{li2019fair, mohri2019agnostic, karimireddy2019scaffold}. \newcite{guha2019one} proposes a one-shot FL algorithm, learning a global model efficiently in just one communication round. \newcite{zhao2018federated}, \newcite{hsu2019measuring} and \newcite{listozec2020federated} study effects of \textsc{FedAvg} and non-i.i.d. client data. \newcite{mcmahan2017learning} and \newcite{hard2018federated} train large recurrent language models with user-level differential privacy guarantees and for mobile keyboard prediction, respectively. \newcite{ge2020fedner} use federated learning for named entity recognition with heterogeneous medical data.

Regarding model size, most architectures used in FL to date are relatively small (e.g., CIFG for mobile keyboard prediction: 1.4M parameters \cite{hard2018federated}), compared to BERT-based language models with hundreds of millions of parameters. How these very large models behave under \textsc{FedAvg} remains underexplored. To the best of our knowledge, \newcite{lin2020ensemble} and \newcite{liu2020federated} are the first ones to train large Transformer models in a federated setting. \newcite{liu2020federated} trained BERT on a medical corpus and showed that both pre-training and fine-tuning could be done in a federated manner with only minor declines in task performance. Nonetheless, the study is mainly a proof-of-concept and does not explore many of the factors that can be expected in real-world scenarios. For instance, the authors only used five clients, and evaluated them only on i.i.d. data. \newcite{lin2020ensemble} introduces FedDF, an ensemble distillation algorithm for model fusion. The authors train a central model through unlabeled data on the client models outputs, and perform fine-tuning on a pre-trained DistilBERT \cite{sanh2019distilbert} in a federated setting as a baseline. To the best of our knowledge, no systematic variation of the number of clients and other relevant factors has previously been explored in this context.

\section{Method}
\subsection{Federated learning}
Federated learning aims to solve the optimization problem
\begin{equation}
\label{fed}
    \min_{\theta\in\mathbb{R}^d} \frac{1}{K} \sum_{k=1}^K F_k(\theta),
\end{equation}
where $ F_k(\theta)= \mathbb{E}_{x\sim \mathcal{D}_k}\left[ \ell_k(\theta; x) \right]$ is the expected loss on client $k$ and $\mathcal{D}_k$ is the data distribution of client $k$. In \textsc{FedAvg}, a global model $f_\theta$ is initialized on a central server and distributed to all $K$ clients, each of which then trains its individual copy of the network using SGD for $E$ local epochs with local batch size $B$.
The clients' updated parameters are then averaged on the central server, weighted by the local data size at each client. The averaged model is distributed to the clients again, and the process is repeated for a defined number of communication rounds.

We implement \textsc{FedAvg} using distributed PyTorch~\cite{paszke2019pytorch}. For each experiment we start from a pre-trained model, and fine-tune it with federated averaging on the current task.

\subsection{Models}
We include BERT with 110M parameters, 12 layers~\cite{devlin2019bert}, ALBERT with 11M parameters, 12 layers~\cite{lan2020albert} and DistilBERT with 65M parameters, 6 layers~\cite{sanh2019distilbert}. This allows us to study the effect that both the parameter count and the number of layers have on \textsc{FedAvg}. All models are the corresponding base models pre-trained on (cased) English. In particular, it should be noted that while the models have similar architectures, they have some key differences. ALBERT introduces factorized embedding parameterization and cross-layer parameter sharing, while the DistilBERT model is a student network trained with knowledge distillation from BERT. We use the weights and implementations of the models available in the Huggingface Transformers library \cite{Wolf2019HuggingFacesTS}.
\subsection{Datasets}

We performed experiments on three standard datasets
to assess the performance of the proposed approach on different tasks.
All of them pose classification problems with a different number of target categories and dataset sizes. For each dataset, we use the test set specified by the source.

\paragraph{IMDB.}
The Large Movie Review Dataset \cite{maasetal2011_imdb} contains of a collection of 50,000 movie reviews and their associated binary sentiment polarity labels  (either ``positive'' or ``negative''), which is used to train a sentiment classifier.

\paragraph{Yelp~F.}
This dataset \cite{zhang2015charlevel} contains reviews of local businesses and their associated rating (1-5). The task is posed as a text classification task, from the review text to its associated rating.

\paragraph{AG News.}
The AG’s corpus of news articles\footnote{\url{http://groups.di.unipi.it/~gulli/AG_corpus_of_news_articles.html}} consists of over one million news articles gathered from more than 2,000 news sources, divided into a number of categories depending on their content. We used the common subset~\cite{zhang2015charlevel} of the whole dataset, consisting of a total of 120,000 samples equally divided in four categories.

\subsection{Experiments and hyperparameters}
We construct several experiments to evaluate how well Federated Learning scales to an exponentially increasing number of clients. In all experiments, the respective dataset is evenly partitioned into a number of subsets equal to the number of clients. Data points are uniformly sampled on each client (i.i.d.) like \cite{mcmahan2017federated}. We do not perform any hyperparameter tuning, and instead, keep all other hyperparameters constant for an unbiased comparison. As baselines we run for each task and BERT-variant a non-federated scenario (clients = 1) with the same configuration.

We run the baselines for a fixed number of rounds based on our compute budget. The test set performance for the baselines are then compared against varying number of participating clients at the same number of rounds. Finally, since runs with a larger number of clients converge more slowly, we allow those runs to continue to a second threshold and report the number of rounds required to reach $90\%$ of the baseline performance, similar to \newcite{mcmahan2017federated}. Runs not reaching $90\%$ of the baseline performance within the second threshold are reported as failures.

We run the baseline for 100 rounds for both IMDB and AG News while setting the second threshold to 200 rounds. However, we only run Yelp~F baselines for 50 rounds due to its large size and set the second threshold at 100 rounds. Like \newcite{lin2020ensemble}, we avoid momentum, weight decay, and dynamic learning rates for simplicity. Instead, all experiments are performed with SGD. Based on \newcite{chi2019classification} we choose the constant learning rate $2 \cdot 10^{-5}$, maximum sequence length $128$ and batch size ($B$) of $32$. Furthermore, the number of local epochs ($E$) is set to $2$ per round.

\section{Results}
\subsection{Fixed compute budget}
\label{sec:results:constant-rounds}
\begin{figure}[ht]
\centering
\includegraphics[width=1\linewidth]{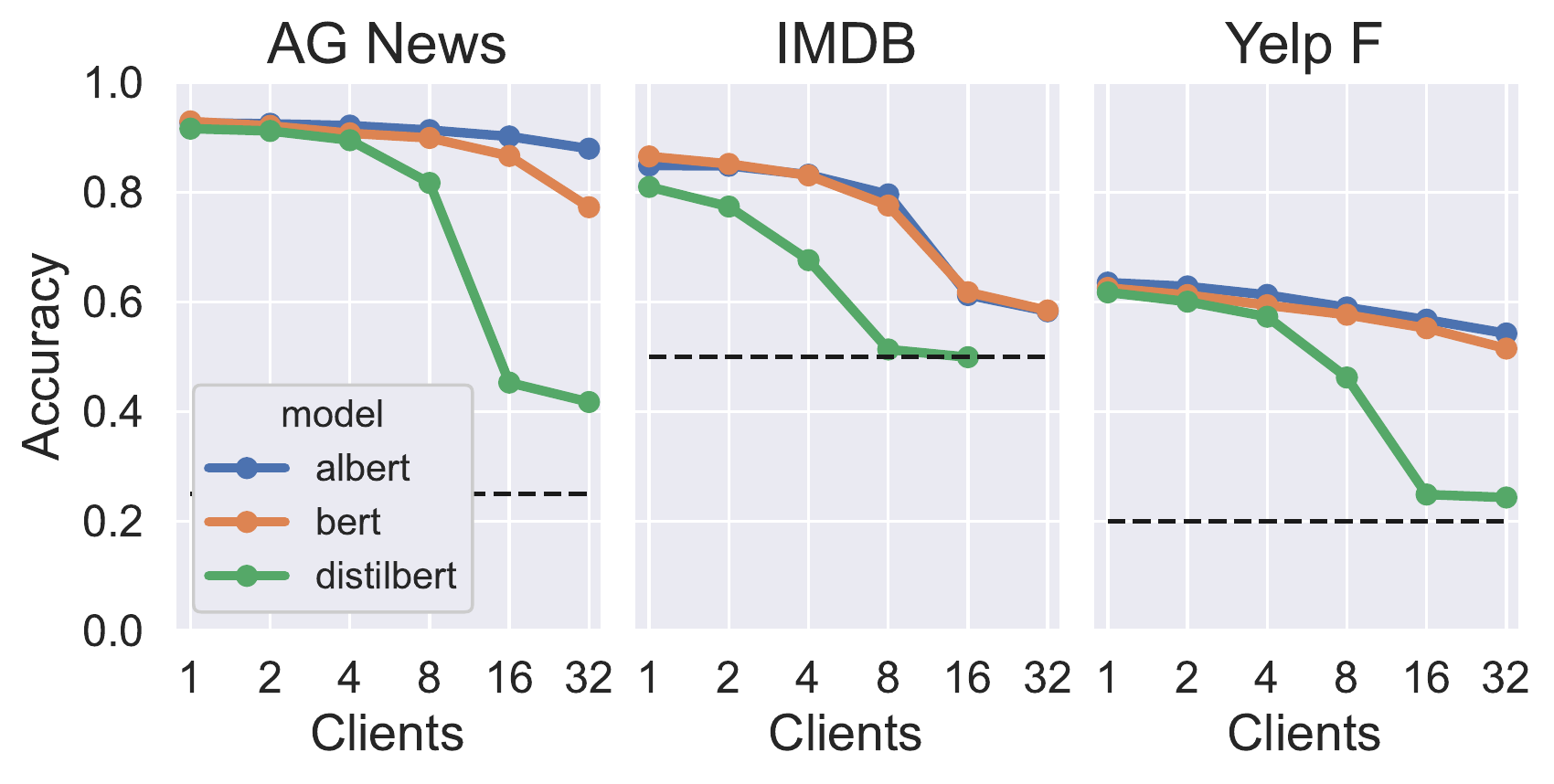}
\caption{Accuracy at a fixed compute budget of 100 rounds for AG, IMDB, and 50 rounds for Yelp~F. The expected accuracy of a random classifier for each task has been highlighted in the dashed line. Higher is better.}
\label{fig:accuracy-distilbert}
\end{figure}
In Figure~\ref{fig:accuracy-distilbert}, we study the effect of increasing the number of clients. It shows the final accuracy after 100 rounds IMDB and AG News, and 50 rounds of the much larger Yelp F., with an exponentially increasing number of clients. Both ALBERT and BERT are well behaved and exhibit a gradual decrease with an increasing number of clients. However, DistilBERT shows a much steeper decline when moving past 4 clients for all datasets, down to the random classifier baseline (IMDB, Yelp F).

\subsection{Rounds until target performance}
\begin{figure}[ht]
\centering
\includegraphics[width=1\linewidth]{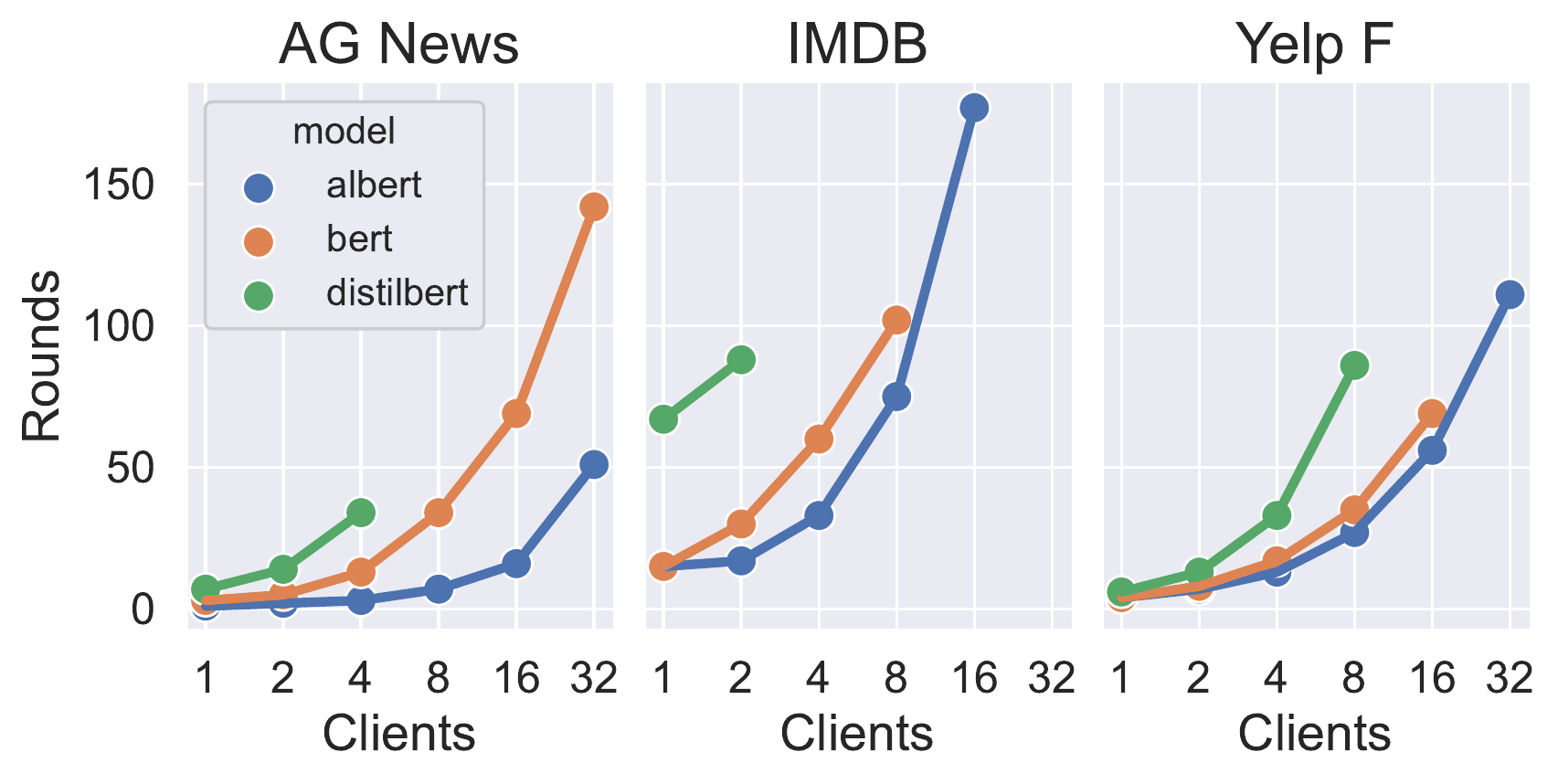}
\caption{Number of training rounds required to reach $90\%$ of the non-federated baseline accuracy. Omittions occur when the target is not reached in 100 (Yelp F) or 200 rounds (AG News, IMDB). Lower is better.}
\label{fig:90percent-rounds}
\end{figure}
Examining the number of rounds necessary to achieve 90\% of the non-federated baseline accuracy (Figure~\ref{fig:90percent-rounds}) yields a similar observation. While all models perform worse with more clients, ALBERT and BERT mostly reach the target accuracy within the allocated number of rounds until 32 clients are used. DistilBERT on the other is unable to reach the target accuracy at 16 clients for Yelp~F, and as low as 4 clients for IMDB.

\subsection{Dynamics of fine-tuning}
\begin{figure}[ht]
\centering
\includegraphics[width=1\linewidth]{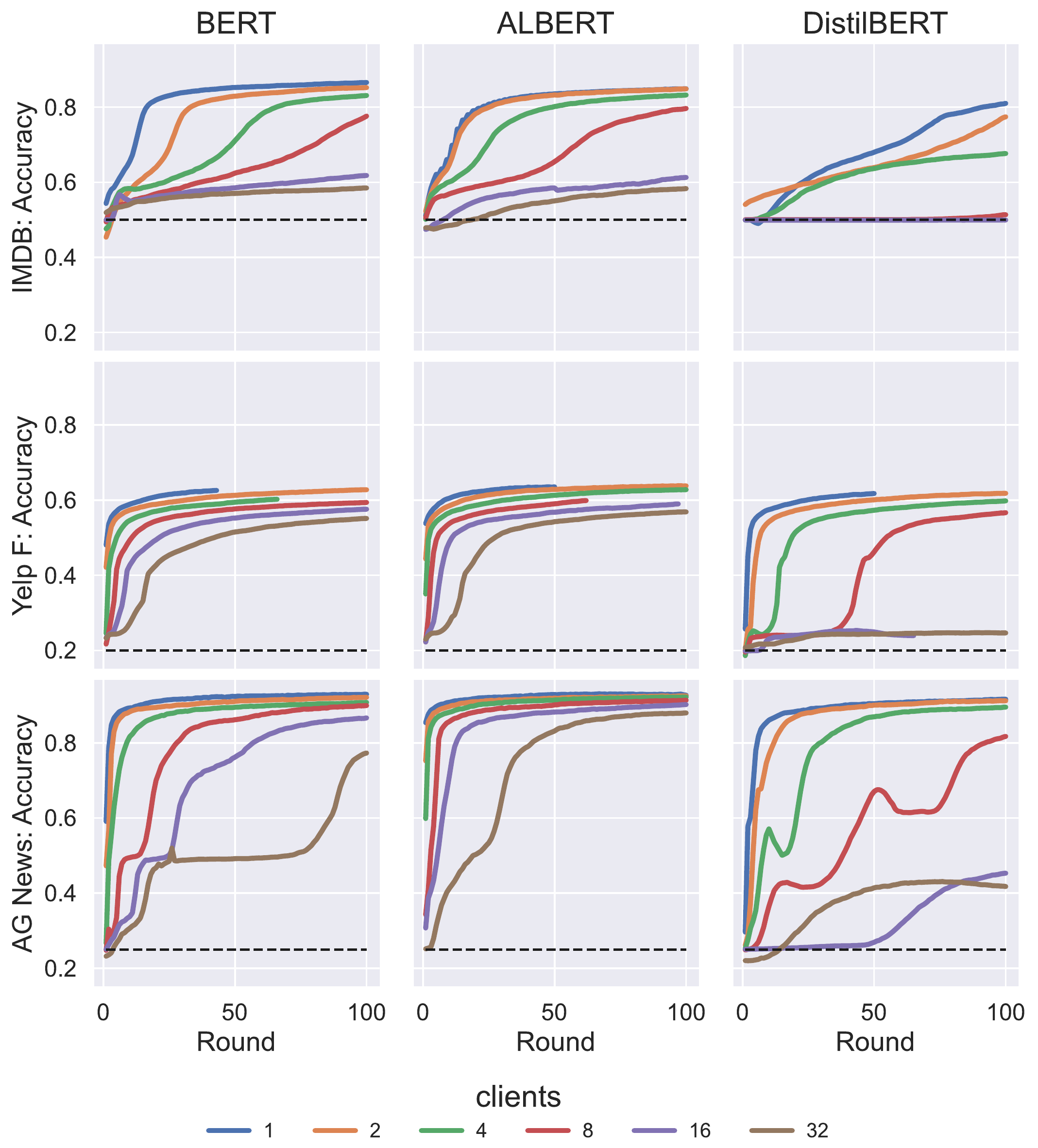}
\caption{Test accuracy (higher is better) over communication rounds for our scenarios. The random classifier baseline is shown as a dashed line.}
\label{fig:learning-trajectories}
\end{figure}

The test accuracy during fine-tuning (Figure~\ref{fig:learning-trajectories}) allows a more complete understanding of how well \textsc{FedAvg} scales for language model fine-tuning. While some scenarios (e.g. Yelp F. with BERT) show a gradual degradation with the number of clients, others configurations are more adversely affected by the increasing number of clients. In some instances the accuracy stays constant over a large period, sometimes even at the random classifier baseline for the whole (DistilBERT on IMDB) or part (DistilBERT on AG News) of the experiment when the number of clients is high.

\section{Discussion}

In this paper, we have evaluated the performance of Transformer-based language models fine-tuned in a federated setting.
While BERT and ALBERT seem to learn each task quickly (Figure~\ref{fig:learning-trajectories}), DistilBERT has a much slower learning progression in the federated setup. A possible explanation is the process of distillation during pre-training, but further research is needed to fully understand why this happens. We demonstrated that BERT and ALBERT scale well up to 32 clients with no sharp decline in performance (Figure \ref{fig:accuracy-distilbert}), but found DistilBERT to struggle at 16 clients in the Yelp~F and AG News tasks, and with as low as 4 clients in the IMDB task, with a substantial drop in performance compared to the baseline. Furthermore, DistilBERT takes more rounds to achieve the same performance. These results demonstrate that we have obtained a higher communication efficiency for BERT and ALBERT as compared to DistilBERT. Further work is required to get a good picture of exactly what affects the communication efficiency for federated learning of Transformer-based language models.

Conversely, the sudden drop in performance in some scenarios indicates that FL can be sensitive to the number of clients. The cause for the instability has not been fully determined. It may be related to both smaller partitions and contradicting models. This highlights the importance of evaluating FL with a varying number of clients at these scales.

For all three models, the performance decline from the baseline is steeper for IMDB compared to the other tasks. This may be related to the variability in the movie review data, adding to a larger inter-client difference in data distribution when data is put into smaller partitions, resulting in a larger difference between the client models taking part in the federated averaging.

In conclusion, we have demonstrated the applicability of the federated learning paradigm and evaluated it on a number of Transformer-based models up to 32 clients. Our findings show that the relatively large sizes of these models are not prohibitive for federated learning.

\section{Acknowledgements}
This work was funded by VINNOVA (grant \mbox{2019-05156}). We would also like to thank \mbox{AI Sweden} and CGit for providing us with compute resources.

\bibliographystyle{acl_natbib}
\bibliography{eacl2021.bib,imdb_wvSent_acl2011.bib}

\end{document}